\documentclass[12pt,epsfig]{article}
\usepackage{epsfig}
\newcommand{\be}{\begin{equation}}
\newcommand{\ee}{\end{equation}}
\newcommand{\bc}{\begin{center}}
\newcommand{\ec}{\end{center}}

\newcommand{\lb}{\label}
\newcommand{\bi}{\begin{itemize}}
\newcommand{\ei}{\end{itemize}}
\begin{document}
\title{Does Meaning Evolve?}
\author{
Mark D. Roberts,\\
     {University of Surrey},  GU2 7XH,\\
            {max1mr@eim.surrey.ac.uk},
{http://cosmology.mth.uct.ac.za/$\sim$roberts}}
\maketitle
\begin{abstract}
A common method of making a theory more understandable,
is by comparing it to another theory which has been better developed.
Radical interpretation is a theory which attempts to explain how
communication has meaning.
Radical interpretation is treated as another time-dependent theory
and compared to the time dependent theory of biological evolution.
The main reason for doing this is to find the nature of the time dependence;
producing analogs between the two theories is a necessary prerequisite to this
and brings up many problems.
Once the nature of the time dependence is better known it might allow the
underlying mechanism to be uncovered.
Several similarities and differences are uncovered,
there appear to be more differences than similarities.
\end{abstract}
{\small\tableofcontents}
\section{Introduction}
\label{sec:intro}
\paragraph{Semantics.}\lb{semantics}
Semantics is the systematic study of meaning, see for example Vendler (1974) \cite{vendler}.
In the $19^{\rm{th}}$ century logic underwent a period of growth.
The main motivation for a renewed interest in logic was a search for
the foundations of mathematics;
however two of the main investigators of the foundations of mathematics,  Frege and Russell,
extended their enquiry into the domain of natural languages.
The influence of mathematical thinking thus
left a permanent mark on the subsequent study of semantics.
To positivists of the Vienna circle such as Carnap
the symbolism of modern logic represented the grammar or syntax of an ''ideal'' language.
The semantics of the positivists ideal language
had been given terms of a relationship connecting
the symbols of this language with observable entities in the world,
or the data of one's sense experience, or both.
Against such a rigid ideal as logic,  natural language appeared to the positivists as vague.
Since a large part of ordinary and philosophical discourse,
particularly that concerning metaphysical and moral issues,
could not be captured by an ideal language,
the positivistic approach provided a way to brand all such talk as nonsensical,
or in some sense meaningless.
The positivists engaged in a prolonged,
and largely unsuccessful,
effort to formulate a criterion of meaningfulness
in terms of empirical verifiability with respect to the sentences formed in natural language.

Another source of dissatisfaction with natural languages is the Sapir-Whorf hypothesis,
see for example Kay and Kempton (1984) \cite{bi:KK}.
This hypothesis of linguistic relativity
implied that the particular language a person learns
and uses determines the framework of his perceptions and thought.
If that language is vague and inaccurate,  as the positivists suggested,
or is burdened with the prejudices and superstitions of an ignorant past,
as some cultural anthropologists averred,
then it is bound to render the user's thinking confused.
Natural language did not remain without champions
in the face of the above two approaches.
The philosophy of "ordinary language" came into its own in the 1940's.
According to the philosophers of this group, natural language,
far from being the crude instrument the positivists alleged it to be,
provides the basic and unavoidable source of all thought;
suggesting that any formal language can make sense only as an extension of,
but never as a replacement for,  natural language.
If philosophical problems arise as a result of a failure
to see the workings of man's language then
they can "dissolve" with improved understanding.
Harris and Chomsky developed transformational, or generative, grammar
thus opening a fresh insight into the syntax of natural languages.
Instead of merely providing a structural description in the form of parsing of sentences,
this approach demonstrates how sentences are built up,
from basic small ingredients,
thus elucidating the formal components of natural language.
\paragraph{Motivation}\lb{mot}
Since Darwin propounded the theory of natural selection in 1859
there have been many attempts to compare and contrast it with other
time-dependent phenomenon.   In fact such endeavours began before then,
for example Monboddo and Tylor's approach to anthropology and also
Spencer's 1852 synthetic philosophy.
Some of the pitfalls and advantages of comparing theories
to biological evolution are illustrated by Cambell and also
Steward's (1956) \cite{bi:steward} and Frank's (1998) \cite{frank}
comparison of biological evolution to cultural change;
this is one of the best known comparisons
of biological evolution to a different theory.
The advantages are that several disconnected societies seem
to go through similar stages,
allowing the possibility of extracting the factors which cause this.
The pitfalls are mainly a result of how to fit interaction between
societies into an evolutionary model.  Societies gather
artifacts and techniques from one another and thus do not
always obtain these in the same order.
The spread of artifacts and techniques can occur
either through
{\it cultural diffusion},
or alternatively the people that posses them can move,
this is know as
{\it demic diffusion}
Renfew (1994) \cite{bi:renfew} p.108.
An example of demic diffusion is the spread of farming through Europe,
Menozzi et al (1978)\cite{bi:menozzi}.
A second frequently made comparison is given by the relationship
between biological evolution and economics,
Alchian (1950) \cite{bi:alchian} p.200:
\begin{quote}
The economic counterparts of genetic heredity,  mutations,  and natural
selection are imitation,  innovation,  and positive profits.
\end{quote}
A third comparison is that ideas from evolution have been applied to
the change in conceptual systems,  Hull (1992) \cite{bi:hull}.
Of course,  most structures change with time and can be referred to as
''evolving'' and of as being a ``process''.   Thus there is the danger of
making straightforward but vacuous remarks when comparing them.
What these examples have in common is an evolving tree structure
(or cladistics),  which can be compared.
In this paper biological
evolution is compared to radical interpretation
There are at least two reasons for doing this
{\it firstly}
is that it allows it to be shown that there
are more differences than similarities between the two theories,
{\it secondly}
is that it allows some aspects of radical interpretation to be
delineated in the more precise terminology of biological evolution.
The relationship between language and biology has also been looked at by
Barbieri (2003) \cite{barbieri}.
\paragraph{Sectional Contents}\lb{sc}
In \S\ref{sec:bioevol} terms from biological evolution are introduced.
The terms which turn out to be the most useful are those of
Gould and Eldridge (1994) \cite{bi:GE} - these terms are not thought by
all biologists to be appropriate for describing biological evolution,
but they have a richer structure which allows more comparisons.
In \S\ref{approaches} {\bf three}, approaches to meaning are introduced.
The {\it first} is the reference approach,
the {\it second} is the truth approach,
which is introduced through a discussion of truth in both formal and natural languages,
the {\it third} is the use approach,
this is similar to the truth approach,
but assigns several attributes in addition to truth to a sentence.
Radical interpretation is concerned with how a language community can
assign truth values to the utterances of other
members of the same community.   Collections of papers on radical
interpretation include Martinich (1996) \cite{bi:martinich}
and Horwich (1994) \cite{bi:horwich};
textbooks include Ramberg (1989) \cite{bi:ramberg}
and Evine (1991) \cite{bi:evine};  and online articles
Kirk (1998) \cite{bi:kirk} and Malpas (1996) \cite{bi:malpas}.
According to the notion of {\sc radical interpretation}
assigning truth is sufficient to assign meaning to these intra-community utterances.
The motivation for this assertion comes from analogy with radical
translation and this in turn is motivated by analogy with truth in
formal languages.   Truth in formal languages is most often described
by Tarski's truth theory.   Meaning in formal languages might be constructed
by the methods of Angluin (1980) \cite{bi:angluin}.   Here truth in natural
and formal languages is discussed in \S\ref{sec:truth}.
In \S\ref{sec:radical} the supposed skill which allows an
anthropologist to understand the language of a previously isolated
community and whether this is a particular example of
{\it radical translation} is discussed.
Traditionally the assignment of truth is what is required
for radical interpretation,  here this is generalized to
allow for the possibility that truth is an inadequate notion
to describe the quality that is assigned to linguistic structure;
thus truth is replaced by the more guarded terminology of
{\it extralinguistic information}.
The extralinguistic information which
drives radical interpretation is usually taken to be supplied by
(the misleadingly termed)
{\it principle of charity};
in \S\ref{sec:coop} this is replaced by the
{\it principle of cooperation dominance}.
In \S\ref{sec:comparison} arguments are given
showing when terms in biological evolution have counterparts in radical
interpretation.
\section{Biological Evolution}
\label{sec:bioevol}
\paragraph{Traditional Evolution}\lb{te}
Biologists account for the origin of existing species from ancestors
unlike current species by invoking the theory of
{\it biological evolution},
see for example Dawkins (1976) \cite{bi:dawkins},
Gould and Eldredge (1993) \cite{bi:GE}.
The evolution of new species is generated by natural selection.
{\sc Natural selection}
is a process resulting in the survival of
those individuals from a population of plants or animals that
were best adapted to the prevailing environmental conditions;
the survivors tend to produce more offspring than those
less well adapted,  resulting in the composition of the
population being changed.
Natural selection is the process of differential reproduction
and has no necessary connection with survival.
{\sc Speciation}
is the origin of new biological species.
Kondrashov (1992) \cite{bi:kondrashov} splits speciation into two kinds:
{\sc allopatric}
speciation occurs when species arise as the by-products
of evolution in geographically isolated populations,
otherwise speciation is
{\sc sympatric}.
{\sc Convergent evolution}
is the evolutionary development of a resemblance between unrelated species.
Although the resemblances between convergent organisms may be superficial
they do not have to be.
Convergence is due to common selection pressures,  not similar environments.
An example is the evolution of wings in birds and insects
because the fore-limbs are homologous as fore-limbs.
{\sc Homologous}
means having the same evolutionary origin,  but sometimes having different functions,
an example is that the wing of a bat and a paddle of a whale are homologous.
In greater generality,
homology is a relation of correspondence
between parts of wholes within the context of a larger whole,
in this case within the context of lineages that share a common ancestry.
Parts are homologous irrespective of function;
but they do not necessarily have different functions.
The anterior appendages of bats and whales make this clear;
they are both locomotory and therefore have the same function of locomotion,
but one is used in flight and the other in swimming
and in that sense they have different functions.
{\sc Analogous}
means having the same function but different evolutionary origin,
an example is that the paddle of a whale and the fin of a fish are analogous.
{\sc Cladistics}
Sneath and Sokal (1973) \cite{bi:SS} p.29
is the study of branching sequences,
contrasted with {\sc phenetics}, which is the study of similarity.
An example is how many branches an evolutionary tree has,
which branch comes off from which other and in what sequence.
If the branches fuse then evolution is referred to as
{\sc reticulate},
Sneath and Sokal (1973) \cite{bi:SS} p.352.
Reticulate evolution is rare in animals
but can occur in plants where
hybridization between distinct species and even distinct genera is possible.
To quote Maynard-Smith (1987) \cite{bi:MS2}, the {\sc Baldwin effect} is
\begin{quote}
If individuals vary genetically in their capacity to learn,
or to adapt developmentally,
then those most able to adapt will leave most descendants,
and the genes responsible will increase in frequency.
\end{quote}
\paragraph{Punctuated Equilibrium}\lb{pe}
Traditionally evolution by natural selection is taken to be both gradual and adaptive.
{\sc Gradualism} means that the changes which occur are
not abrupt and happen in small stages.
An {\sc adaptation} is property of the component organisms of a species that
has been acquired because its presence has favoured the survival
and reproduction of those organisms that possessed it.
Adaptionistic means changes which happen in this manner.
These days both of these properties are in contention.
Instead of evolution proceeding by gradualism it can proceed by punctuated
equilibrium as discussed by Gould and Eldredge (1993) \cite{bi:GE}
and by Pinker and Bloom (1990) \cite{bi:PB} p.711.
Pinker and Bloom describe
{\sc punctuated equilibrium}
as follows
\begin{quote}
According to the theory of punctuated equilibrium most evolutionary change
does not occur continuously within a lineage,
but is confines to bursts of change that are relatively brief on
the geological time scale,  generally corresponding to speciation events,
followed by periods of
{\sc stasis}.
\end{quote}
Instead of adaptionism there might be indirect and implicit natural
selection through exaption and spandrelization.
{\sc Exaption},  formely {\it preadaption},
means the adoption of a character that had one use in an ancestral form in to a new,
different use in a descendent form.
For example,  the bones in the jaws of the ancestors of mamimals were exapted into the
hammer,  stirrup and anvil,  the bones of the middle ear.
{\sc Spandrelization}
means uses for parts which originally occurred as a by-product necessitated by other structure.
The idea of spandrels has been subjected to considerable criticism.
Spandrels come in at least two types.
{\sc Ephenomenal spandrels}
in which the structure is simply a by-product,
examples are:  the redness of blood,  the hollow at the back of the knee.
Spandrels which have since been changed by direct
natural selection are
{\sc modified spandrels},
examples of these are chins and hexagonal honeycombs.
Bowring et al (1993) \cite{bi:bowring} have calibrated
the rate of early Cambrian evolution and found that the Tommatian and
Atdababian stages lasted less than ten million years.
During these stages a large variety of new life forms appeared,
therefore biological evolution must have scope to accommodate rapid
change - traditional gradualism cannot do this.
\section{Approaches to Meaning.}\label{approaches}
\paragraph{Meaning and Reference.}\label{mandr}
The problem of meaning and reference can be approached through the following steps,
see for example Vendler \cite{vendler} (1974).
The perception of physical entities such as objects
might lead an intelligent being to the thought of a related happening with some regularity.
For example, the sight of smoke evokes the idea of fire.
The smoke is thus a sign of some related happening.
It is a natural sign,
because the connection between the sign and the thing signified is a causal link.
The connection between the symbol
and the thing signified in cases such as road signs is not a natural one;
it is established by tradition or convention and is learned from these sources.
Non-natural signs or symbols are widely used in human communication.
The elements of language appear to be non-natural signs.
The interest in words and phrases reaches beyond their physical sound as
their perception is likely to direct attention or thought to some related happening.
Words are the main media of human communication,
and as the diversity of languages shows,
the link involved between words and what they signify cannot be in a simple correspondence.
Words mean things that they make us think of;
the {\it meaning} of the word can be thought of as the relationship
that connects it with that thing,  this is sometimes called correspondence,  see \S\ref{cn}.
There are some words for which this approach seems to work.
The name Paris can refer to the city of Paris.
Beginning with Plato the initial plausibility of such examples
created an obsession in the minds of many thinkers.
Regarding proper names as excellent examples of words,
they tried to extend this {\it referential} model of meaning
to all of the other classes of words and phrases.
Plato's theory of 'forms' can be viewed
as an attempt to find a referent
for such common nouns as 'cat'
or for abstract nouns like 'beauty'.
The word Socrates in the sentence
'Socrates is wise' refers to Socrates suggesting that
the word wise refers to the form of wisdom.
Unfortunately whereas Socrates was a real person in this world,
the form of wisdom is not something that has been encountered.
The difficulty representing by 'platonic' entities
of this kind increases as one tries
to find appropriate referents for verbs, prepositions, connectives, and so on\dots.

There are at least {\bf two} more serious problems with the referential theory of meaning.
The {\it first} pointed out by Frege,
is that two expressions may have the same referent
without having the same meaning.
His example is that ''the Morning Star'' and ''the Evening Star'' denote the same planet,
Venus,  however the two phrases do not have the same meaning.
If they had, then the identity of the Morning star and Evening star
would be as obvious to anybody who understands these phrases.
The identity of the Morning Star with the Evening star
is a scientific not a linguistic matter.
Thus in the case of names,
it is necessary to distinguish between the referent of a name,
and its connotation or its meaning.
The {\it second} problem with the theory of referential meaning
arises from meaningful phrases that pretend to refer to something but do not.
A well-used example
in the case of such a definite description is ''the present king of France'',
the phrase is meaningful although there is no such person.
If the phrase were not meaningful,
one would not even know that the phrase has no actual referent.
Russell and Quine's analysis of these phrases,
detached meaning from reference by claiming that these expressions,
when used in sentences,
are equivalent to a set of existential propositions which are
propositions that do not define reference.
The example ''the present king of France is bald''
comes out as
"there is at least, and at most, one person that rules over France,
and who ever rules France is bald".
These propositions are meaningful, true or false, without definite reference,
thus the hope that meaning could be understood in terms of reference is false.
\paragraph{Truth in Formal Languages}\lb{sec:truth}\lb{tfl}
The positivists suggested verification as the criterion of empirical meaning;
this is to understand a sentence is to know what state of affairs would make it true or false.
Verification suggests a truth theory of meaning.
Here how truth is defined in formal languages is considered
before how to apply this to natural languages.
At first sight it would seem that if the truth or otherwise of a
sentence is known then by necessity its meaning must be known;
furthermore the meaning of a sentence can be known but its
truth be undetermined.   Radical interpretation reverses
this picture and takes truth as the primary notion.
The truth value of a sentence is taken to be assigned by appealing to
extralinguistic information,  allowing the meaning of a sentence
to follow in a manner which allows the
best fit to these truth values.   How the assigning of truth values is
accomplished is discussed in the next two sections,  this section is limited
to conceptions of truth.
Theories of truth have been comprehensively introduced in Kirkham (1995) \cite{bi:kirkham}.

Tarski (1967) \cite{bi:tarski} p.63 and Evine (1991) \cite{bi:evine} p.82
point out that Aristotle had a notion of truth, expressed in his Metaphysics
\begin{quote}
To say of what is that it is not,  or what is not that it is,  is false;
while to say of what is that it is,  or what is not that it is not,  is true.
\end{quote}
Tarski's approach is to
take a formal object-language which has sentences that can be true or false,
and then compare these sentences to sentences in a meta-language.
In order to consider what would be involved in a definition of truth,
consider what would be involved in any definition.
There are two
formal conditions of adequacy
and one material condition of adequacy:
\begin{description}
\item [formal condition of adequacy 1]
the definition must take the form of a one place predicate,
\item [formal condition of adequacy 2]
the {\sc definiens} must not contain the\\
{\sc definiendum} or words constructed from the {\sc definiendum},
\item [material condition of adequacy]
the definition must entail all and only instances of the word being defined.
\end{description}
The condition
of material adequacy for a definition of truth can be expressed formally using
{\it criteria}
$W$ (also called convention T,  form T,  schema T):
\be
{\rm (W)}~~~~~~~~s\varepsilon T~{\rm iff}~p
\lb{eq:W}
\ee
where $s$ is sentence in the object-language,
and $p$ is a sentence in the meta-language.
A sentence for which ($W$) holds is called a
$W$-{\it sentence}
(also a T-sentence,   Evine (1991) \cite{bi:evine} p.80).
The sentence in the object language are said to be true iff (if and only if)
they are true in a meta-language.  The same approach works for languages
which have any finite number of sentences.

Now consider a formal object language which consists of a finite
number of sentences and the connectives
$\{not,~  and,~  or,~  if~ \ldots~ then \}$ of the predicate calculus.
With these one can construct sentences of infinite
length and also an infinite number of sentences.   One says a sentence is
true iff it is a combination of true sentences and connectives which are
well formed formulas (wffs) and are true according to the rules of the
predicate calculus.

Next one needs a conception of truth which will work for quantified
predicate logic.   Quantifiers are $\forall$  (for all)
and $\exists$ (there exists),
and these range over the variables in a sentence.   A variable is free in a
wff if it is not bound by a quantifier.   A closed sentence has no free
variables,  every word is specified.   A well formed sentence is open if
it has an unbound variable.   An example is '$x$ is the father of John',
it is open because $x$ is not specified.   Truth cannot be assigned directly to
open sentences,  one has to introduce the notion of satisfaction.   First
consider open sentences with one free variable,  then the sentence is
{\sc satisfied}
if there is a particular choice of free variable for which it is true.
Sentences can be further complicated by having more than one
unbound variable,  there can be an infinite number.
In this case an open sentence is satisfied if there is a
{\it sequence}
(ordered set) for which it is true.
A {\sc true sentence}
in the object language is a sentence which is
satisfied by every sequence.
\paragraph{Truth in Natural Languages}\lb{tnl}
It is necessary to have a view on what is true or otherwise in natural languages in order to
discuss radical translation.  Initially definitions of
truth which worked in the context of formal logic were found,
and then it was debated whether or not they could be extended to natural languages.
Usually meaning is a more basic property of both sentences and communication than truth.
One cannot ascertain whether a sentence is true or false if one does not know what it means.
In radical translation this is reversed,  it is assumed that one knows what sort of
communication is true or false,  and then from this constructs an idea of meaning.
This process of construction being called radical translation.
Tarski (1969) \cite{bi:tarski} p.65
thinks that the formal language definition of truth cannot be extended to
natural languages because of the antinomy of the liar.
I agree with this,
this is one of the reasons that I have for reformulating radical
interpretation with extralinguistic information assigning a quantity $Q$ to
sentences,  see Roberts (2003) \cite{mdr34}.
There are other problems with truth in natural languages,
such as sentences often contain demonstratives.
{\sc Demonstrative}
means denoting or belonging
to a class of determiners used to point out the individual referent or
referents intended,  examples are: this,  that,  these,  and those.
Sentences which contain demonstratives in isolation are not
indexed to a particular context,  especially a specific place and time;
because of this such sentences are sometimes referred to as
{\sc indexical}.
Thu the problem arises as to how to assign a truth value to a single sentence
which contains an indexical.
Davidson's (1984) \cite{bi:davidson2} p.131 approach is this
requires a new theory of truth,  which incorporates natural languages and
can be used as a theory of interpretation,  and thence meaning.
In Davidson's approach the first
problem that needs to be addressed is whether Tarski was correct in his
contention that his definition of truth cannot be extended to natural
languages.   Davidson (1984) \cite{bi:davidson2} p.133 suggests
splitting the problem of truth in natural languages into two stages:
\begin{quote}
In the first stage,  truth will be characterized,  not for the whole language,
but for a carefully gerrymandered part of the language.   This part,
thought no doubt clumsy grammatically,  will contain an infinity of sentences
which exhaust the expressive part of the whole language.   The second part
will match each of the remaining sentences to one or (in the case of ambiguity)
more than one of the sentences for which truth has been characterized.
We may think of the sentence to which the first stage of the theory applies
as giving the logical form,  or deep structure, of all sentences.
\end{quote}
I think what Davidson means is that radical translation is a multi-stage process.
In the first stage one has a simple part or segment of language,
which would say consist of just yes and no assertions in active voice and so on.
To this simple subset of language truth and then meaning are assigned.
Then the subset of language is enlarged and the process starts over again,
until finally most of the language is encompassed.
I am not sure if all of the language is ever reached,
after all ambiguity is an aspect of more complex communication.
My view is that the first Davidson stage requires two incompatible criteria to be met
simultaneously,  this stage is required to both
1)exhausts the expressive power of the language,  and
2)is only a gerrymandered part of it;
and that the second stage is redundant - once the sentences
have been matched they will have logical form and hence be part of the first stage.
An extreme view is radical pragmatism,  the proponents of which hold
that there is no such thing as literal meaning,  the implications of this is
that there is no such thing as a theory of truth even for a formal language.

Another problem with assigning logical form to sentences is that in the
empirical sciences results of experiments are analyzed statistically,
resulting in statements being probabilistic:  however for formal languages
of the type considered by Tarski predicate logics etc\dots are used.
Yet another problem is that in formal logic 'implies' has a definite meaning.
The problem is that 'implies' also has well defined meaning elsewhere,
specifically in probability theory,  and the two definitions are not entirely compatible.
So the question arises,  why not have a theory of meaning which has probability
theory as its starting point rather than formal logic?
This would initially entail assigning probabilities to sentences rather than just yes or no.
The way I think of this,  is to keep the nature of what is assigned open and call it $Q$,
see Roberts (2003) \cite{mdr34}.
The different properties of 'implies' are discussed in Calabrese (1987) \cite{bi:calabrese}.

Many utterances of natural language are not true or false at all.
Whereas statements, testimonies, and reports are true or false;
orders, promises, laws, regulations, proposals, prayers, curses,
and so on are not assessed in terms of truth or falsity.
The employment of words in these speech acts is not necessarily less relevant to their meaning
than their use in speech acts of the truth-bearing kind.
\paragraph{Meaning and Use.}\label{mandu}
The difficulties just mentioned lead to another
theory of meaning sometimes called {\em use theory},
see for example Vendler \cite{vendler} (1974).
This view admits that not all words refer to something,
and not all utterances are true or false.
What is common to most words and sentences,  is that people use them in speech.
Consequently their meaning could be nothing more than the rules
that govern their employment,
or in other words how they are used
Use theory originated from at least {\bf two} places.
{\it Firstly}
in trying to understand the nature of moral and aesthetic
discourse it was sometimes suggested that words such as 'beautiful'
have an emotive meaning instead of, or in addition to,
the descriptive meaning other words have;
understanding the role or use of such words is to know their meaning.
{\it Secondly}
Wittgenstein's (1958) \cite{bi:wittgenstein}
language games in which uses that do not convey truth plays a dominant role.
Human language is apparently a combination of a great many language 'games'.
'Games' is how Wittgenstein refereed to a segment of language where a set of given
rules are applied,  it is an unfortunate nomenclature as it suggests frivolity,
however it is now firmly established.
The principle of meaning according to use theory is:
the meaning of a word is the function of its employment in these games.
To Wittgenstein the question 'What is a word really?'
is analogous to 'What is a piece in chess?'
This led Austin to offer a systematic classification of the variety of speech acts.
Define {\sc illocutionary} as
an act performed by a speaker by virtue of uttering certain words,
as for example the acts of promising or of threatening.
Define {\sc performative} as
denoting an utterance that itself constitutes the act described by the verb.
For example,  the sentence {\it I confess that I was there} is itself a confession.
Define {\sc perlocution} as
the effect that someone has by uttering certain words,
such as frightening a person.
According to Austin, to say something is to do something,
and what one does in saying something is typically indicated
by a particular performative verb prefixing a "normal form" of the utterance.
These verbs, such as 'state',  'declare',  'judge', and so on,
mark the Fregean force of the utterance in question.
If one says 'I shall be there' then depending on the circumstances,
this utterance may amount to a prediction, a promise, or a warning.
The dimension of truth and falsity is not invoked
by all the utterances of the language;
therefore it cannot provide an exclusive source of meaning.
In other words truth is one of several attributes.
Other attributes could be feasibility,  utility,  and moral worth.
These attributes could be as much involved in the understanding of
a sentences as the attribute of truth,
and perhaps can be identified with the quality $Q$,
see Roberts (2003) \cite{mdr34}.
\section{Radical Translation and Interpretation}
\label{sec:radical}
\paragraph{Radical Translation}\lb{rt}
{\sc Radical translation},
Quine (1960) \cite{bi:quine1} Ch.2,
is a hypothetical skill in which an `anthropologist' attempts to assemble
a translation manual for an unknown language.
Quine gives an example of how the translation manual could be compiled,
it involves the anthropologist noting that a `native'
utters the word {\it gavagai} when a rabbit runs past.
The anthropologist then sets up a hypothesis to test,
he guesses that '{\it gavagai}' means 'there is a rabbit';
he then notes in future when '{\it gavagai}' means 'there is a rabbit'
match up,  in order to see if any qualification to the translation is
necessary and to confirm that the translation has been successful.
In this behaviourist manner a translation manual between the two languages evolves.
\paragraph{Radical Interpretation}\lb{ri}
Radical interpretation is a supposed skill similar to radical translation,
but where there is only one language present.
Meaning is assigned to sentences in the language when
extralinguistic evidence suggests that it is appropriate to do so;
the nature of the extralinguistic evidence is not made explicit;
for example there is little variety in the emotional impact of
facial expressions,  Ekman {\it et al} (1983) \cite{bi:ekman},
so that it is difficult to imagine substantial extralinguistic
evidence originating there.
Lewis (1974) \cite{bi:lewis2} takes the position that radical interpretation
applies to all aspects of an individuals behaviour;
this would include intentions and beliefs:  meaning is sometimes expressed
in terms of these Grice (1989) \cite{bi:grice} p.96.
Lewis (1974) \cite{bi:lewis2} restricts himself to interpretation of an individual,
although there appears to be no bar to applying it to a community.
In this case interpretation would work not only for sentences
but for all socially contrived structures:
in particular memes and conventions,  defined in the next paragraph,
would be subject to interpretation.
There is a danger of having a circular chain of reasoning here,
what if the assignment of extralinguistic information
is itself subject to interpretation.   Provided such `self-interaction'
is small it should not interfere with interpretation.
\paragraph{The Size of Structure Interpreted:Memes,Glossemes, and Conventions}\lb{ss}
Gesture and speech have been noted to be used in parallel in human communication,
McNeill (1985) \cite{bi:mcneill},
and the question arises as to how large
a chunk of human activity this assigning meaning applies to.
Three contenders are:
\bi
\item
{\it glossemes},  Bloomfield (1926) \cite{bi:bloomfield},
The smallest meaningful units of language are called {\sc glossemes}.
Most glossemes have an associated meaning,
this is called a {\sc noeme}.
\item
{\it memes} Dawkins (1976) \cite{bi:dawkins}.
A simple definition of a
{\sc meme},
is an artificially created,  non-physical,  replicating structure.
\item
{\it conventions} Lewis (1969) \cite{bi:lewis1},
Particular cases of memes are given by {\sc conventions},
a simple definition of this is a rational self-perpetuating regularity.
\ei
In the definition of a meme artificial means man-made;
and non-physical means that it is an 'idea' rather than something more readily tangible,
here questions in the philosophy of mind such as whether an idea is really non-physical
are not addressed.
A computer virus can be argued to be either physical or non-physical,
physical seems more plausible as it has a physical presence which
in practise is often identified,
thus a computer virus is not a meme.
The original definition of meme \cite{bi:dawkins} was as a unit of intellectual or cultural
information that survives long enough to be recognized as such,  and which
can pass from mind to mind.
It has led to the theory of memetics,  see  Dennett (1995) \cite{dennett}.
The extent of a meme is nebulous,  for example it can
consist of single note in a symphony or to the idea of symphonies.
The reason for introducing memes is to emphasize the analogy between
the continuity of ideas held by a language community.
This analogy can also be extended to virus,
apparently an example of a 'mind virus' are chain letters,
Goodenough and Dawkins (1994) \cite{bi:GD}.
A meme and its perpetuation are not explicitly physical.
In the context of conventions 'rational'
is a limitation on what can be considered a convention,
it is a convention which side of the road traffic drives on,
however if traffic drives on both simultaneously this is not a convention.
\section{Cooperation Dominance}
\label{sec:coop}
\paragraph{The Origin of Cooperation Dominance}\lb{mp}
Radical interpretation depends crucially on there being a mechanism by which
extralinguistic information can be used to give truth values to sentences.
To discuss this begin by noting that in radical translation the natives
might deliberately mislead the anthropologist for a variety of motives.
For the translation to work deliberately misleading
occasions must be in the minority or be detectable by the anthropologist.
This observation is elevated to a principle,  here called
{\sc cooperation dominance},
which can be simply defined as
the disposition of people to earnestly cooperate preponderating
over alternative modes of behaviour.
Traditionally,   linguistic cooperation dominance has been referred to as the
{\sc principle of charity}
and it has been thought
necessary to invoke it so that sentences can be given truth values.
Quine (1974) \cite{bi:quine2} p.328 attributed the phase
principle of charity to N.Wilson.
Lewis (1974) \cite{bi:lewis2} p.338 holds that it is a collection of
several sub-principles.
The principle of cooperation dominance ensures that cooperation predominates,
this essentially `{\it drives}',
(alternatively `optimizes the process of' or `forces coherence of')
radical interpretation and results in meaningful language.
The accuracy and speed at which this occurs is positively
correlated with fitness for survival thus ultimately radical interpretation
is driven by biological evolution.
The actual mechanism by which radical interpretation occurs
might be autocatalytic,  Roberts (1999) \cite{bi:mdr99}.
\paragraph{Ploy and Counter-Ploy}\lb{cp}
Radical translation is one of at least thirteen related topics involving the interplay of
deceit and cooperation and its evaluation.
No detailed comparison is given here.
To list the thirteen in no special order:
\begin{enumerate}
\item
{\it radical translation}
as discussed in the previous paragraph,
\item
accounting for
{\it coterie language}.
An example is Cockney rhyming slang.
This apparently developed to confuse immigrant workers about the intentions
of native (Cockney) workers in the east-end docks,
\item
evaluation the
{\it credibility of communication}.
Social psychologists Ravin and Rubin (1976/83) \cite{bi:RR} p.176,
study the credibility of communication,  which entails gauging how useful and
truthful communication is,
\item
accounting for
{\it animal deceit}.
The collection of papers in Byrne and Whiten (1988) \cite{bi:BW}
discuss devious behaviour by animals and together these make a good case
that the reason human intellect evolved,  was in order to
enhance social manipulation and expertise in preference to the
prosaic explanation of greater ability to collect food and avoid predators,
\item
accounting for
{\it human cognitive abilities}.
As pointed out by Pinker and Bloom (1990) \cite{bi:PB} p.724,
the possession of language is an enormous
advantage that biological evolution will select for strongly.
The reason is that it allows
communication of a large amount of information which aids
survival - this leaves open the nature of the useful information.
Survival can be enhanced sometimes by cooperation and
sometimes by selfishness and the interplay of these two factors as great
impact leading to a ``cognitive arms race''
Pinker and Bloom (1990) \cite{bi:PB} p.725,
\item
accounting for {\it irrationality}.
On occasion people do seem to make decision that are disadvantageous to
themselves,  Kahnemann {\it et al} (1982) \cite{bi:kahnemann},
to reconcile this with the principle of cooperation dominance it is
necessary to assume that people behave rationally with greater frequency.
A reason for this could be that the experiments of Kahnemann {\it et al} isolate
context in a way that is only discernible to the patient;
at first sight it might not be rational to persevere with the contrived
nature of the experiment,
\item
{\it nonconsequentialist decisions},  Baron (1994) \cite{bi:baron}
are made by some people regardless of their judgement of what the consequences
might be,
\item
accounting for {\it stupidity}.
Attempts have been made to define stupidity Welles (1984) \cite{bi:welles},
his definition requires nonrational behaviour,
this occasionally has survival advantage;
the reason for this is that it enhances social cohesion in a peer group,
see also de Botton (2004) \cite{botton}.
Non-rational behaviour has also been discussed by
Mackay (1852) \cite{mackay},
Shermer (1997) \cite{bi:shermer},
and Wheen (2004) \cite{wheen}.
Group thinking can sometimes have beneficial effects and sometimes detrimental effects,
Janis (1982) \cite{janis},  and Surowiecki (2004) \cite{surowiecki}.
It seems that when the need to preserve group harmony and cohesiveness preponderates
it is detrimental and when there is some self-interest involved it is beneficial,
\item
using {\it game theory} to explain behaviour.
The interplay of cooperation and deviousness
might be quantifiable using game theoretic methods:
game theory has been used in the behavioural sciences for example by
Maynard-Smith (1984) \cite{bi:MS},
and also by Glance and Huberman (1994) \cite{bi:GH} who find that the
amount of cooperation is inversely proportional to group size,
\item
evaluating {\it relevance},  see Sperber and Wilson (1987) \cite{bi:SW}.
As argued in the fifth point communication allows transfer of large
amount of useful information:  for it to be useful suggests that it is in
some way relevant.
From this point of view the principle of cooperation dominance is
just a generalization and extension of the {\it maxim of relevance}
entertained by Grice (1989) \cite{bi:grice} p.27,  as relevance ensures
speedy communication of information,
\item
various {\it biological altruistic} approaches,
Hamilton (1964) \cite{bi:hamilton} and Trivers (1971) \cite{bi:trivers},
\item
the study of cooperation in small scale societies Henrich {\it et al} \cite{henrich} (2001),
\item
{\it equity theory},  Walster {\it et al} \cite{WWB} (1978) which requires the following:\\
1.    In an interpersonal relationship,
a person will try to maximize his or her own outcomes (where outcomes =
rewards - costs).\\
2a.    By developing systems whereby resources can be equitably
distributed among members, groups can maximize
the probability of equitable behavior among members.\\
2b.    A group will reward members who behave equitably
toward others and punish those who do not.\\
3.    Inequitable relationships are stressful for those within them.
The greater the inequity, the greater the distress.\\
4.    A person in an inequitable relationship will take steps
to reduce the distress by restoring equity.  The more distress
felt, the harder the person will try to restore equity.
\end{enumerate}
\section{Comparison of the Theories}
\label{sec:comparison}
\paragraph{The Counterpart to Adaptation by Natural Selection.}\lb{cn}
A {\sc correspondence theory},
Ramberg (1989) \cite{bi:ramberg} Ch.4,
is a theory that attempts to elucidate a relation between language
and the world:  the meanings of the language are called
{\sc intensions},
and the elements of the world are called
{\sc extensions}.
Now assume that a correspondence theory of language is correct for
a significant segment of natural language.
In both biological evolution and radical interpretation there is an
optimization process occurring,
in biological evolution it is the optimization of a species to fit
its environment through natural selection,
in radical interpretation it is the optimization of the correspondence between
intensions and extensions.
Thus just as organisms which are not fit are selected against,
correspondences which are inaccurate are selected against.
The counterparts to intensions and extensions could be
intensions corresponding to genotype and extensions corresponding to phenotype,
but this does not seem to fit.
In radical interpretation increasing correspondence is not the only optimization process,
there is also
{\sc coherence}
which means the tendency of meanings assigned to a segment of
language to develop in a way which increases self-consistency.
If the manner of development is by selecting self-consistent meanings then
coherence appears to have the idea of selection already contained in its definition.
Another way of expressing coherence is that an individual's understanding
of a given part of a language develops in such a way that what it means
tends to have less internal contradictions.
The method by which such development could occur would involve selection
amongst various hypotheses concerning what the part of the language means.
Having two optimization process in radical interpretation is another
difference from the single process in biological evolution.
At this point it should be emphasized that we are attempting to explain
language meaning as opposed to linguistic diversity,
Renfew (1994) \cite{bi:renfew} and Hurford (1998) \cite{bi:hurford}.
Linguistic diversity is concerned with how specific language structure
spreads,  geographically or otherwise.
Language meaning is concerned with how a segment of language carries meaning.
Perhaps the two notions are related and furthermore it might be
possible to quantify by how much.
This might be done by appealing to colour terms,
Berlin and Kay (1969) \cite{bi:BK}.
Assume the description of the eleven focal colours to be given by eleven
noemes (defined in \S\ref{ss}),
now the spread of the noeme and the corresponding word can be compared.
To summarize the counterpart of biological natural selection
is the increase of accuracy of both correspondence and coherence.
\paragraph{Mutations}\lb{mu}
Mutations often involve duplications, translocations,  and chromosomal rearrangements.
Chromosomal rearrangements in biology have counterparts in language;
the straightforward counterparts are given by words that have occurred
through speech errors (such as blends) later being
adopted by the language community.
A {\sc blend} also called a {\it portmanteau word},
is a word formed by joining together the beginning and end of two other words,
an example is 'brunch' which is a blend of 'breakfast' and 'lunch'.
Mutations account for biological diversity but not all,
and perhaps not even a significant part of,
language diversity can be accounted for by appealing to speech errors.
Language is creative and at first sight it is hard to accommodate
all of this creativity by appealing to mutations.
It may well be that creativity can be accounted for by
having a large number of mutant ideas and then discarding
the majority of them,  but this would be hard to verify.
By mutant ideas I mean ideas which are original,
but do not necessarily posses any other attribute,
they may well not even be well-defined.
To summarize mutations do occur in language,  but that the content
of the language mainly originated by creativity,
whether creativity can be accounted for by mutations is an open question.
\paragraph{The Smoothness of Change}\lb{sm}
In biological evolution change can be gradual or proceed by punctuated equilibria.
The question arises as to whether an interpreter's interpretation
of a language changes gradually or via punctuated equilibria.
How large a segment of language radical interpretation applies to,
has been discussed at the end of \S\ref{ss};
for simplicity now interpretation is restricted to a single individual.
Punctuated equilibria applies to populations rather than individuals,
the punctuated equilibria population corresponds to the population
of parts of a language under interpretation in a single individual.
Consideration of whether an interpreter's interpretation changes gradually
or by punctuated equilibria depends crucially on the time-scale involved:
here long time scales (of the order of years) are called
{\sc coarse-grained},
and short time scales (of the order of seconds) are called
{\sc fine-grained}.
{\bf First} of all consider this question on the fine-grained scale.
Radical interpretation proceeds by assigning truth
values to sentences,  consider the input of a single sentence and
the co-occurring assignment of a single truth value.
Before the utterance there would be one
{\sc fleeting theory}
of the language and after it another;  thus the theory of the language
jumps discontinuously by a discreet amount:
hence on the fine-grained scale the theory proceeds via punctuated equilibria.
This conclusion is dependent on the extralinguistic information assigned
being a two-valued (or bi-valent) quantity.
Discontinuous jumps will also occur if the extralinguistic
information comes in integer valued packets;
however if it is a real-valued quantity (or quantities)
then continuous gradual change might be possible.
Davidson (1984) \cite{bi:davidson2} p.279 says:
\begin{quote}
No doubt we normally count the ability to shift ground appropriately
as part of what we call 'knowing the language'.
\end{quote}
From the fine-grained perspective the skill in making small
adjustments and shifting ground appropriately would be the same as knowing the language,
however there is no existent psychological mechanism to demonstrate how this is done,
or how it might manifest itself in consciousness.
{\bf Secondly} on the coarse-grained time scale
an interpreter is said to hold a
{\sc quiescent theory}
of a language.
This quiescent theory needs occasional polishing up and rarely substantial
adjustments;  the way the quiescent theory changes can be illustrated by
the  way that single  words change meaning - a new word can be invented,
in which case this part of the theory has a punctuated equilibrium,
however this has just a small impact on a small section of the theory;
hence the quiescent theory usually proceeds by gradualism.
In the Elizabethan era Shakespeare,  Jonson,  and others added
thousands of new words to the English language;
as this is on the coarse-grained time scale it effects the quiescent theory.
Whether this is an example of a quiescent punctuated equilibria depends on
the quantity of change thought to be necessary for such a classification.
Previous writers,  e.g. Ramberg (1989) \cite{bi:ramberg} p.102,  hold that
an interpreter at a given moment holds a
{\sc passing theory}
of the language.
The above Davidson quote suggests to me that a passing theory
is the same as what I have called a fleeting theory,
and furthermore that this is adjusted every time that an utterance is heard,
{\it i.e.}  radical interpretation of the fleeting theory is an actively
occurring process in normal speech.

A similar distinction has been made in language acquisition
Campbell (1979) \cite{bi:campbell}.
Several time scale terms are used here.
The first is the
{\sc microgenesis}
of the language,  which is the rapid,
moment-by-moment nature of everyday language processes,  by virtue of which we
understand and produce utterances on a time scale that is marked off in
seconds and milliseconds.
In contrast there is the
{\sc macrogenesis}
of language,
this can either refer to an individuals learning of a first or subsequent
language (the
{\sc ontogenesis}
of language) on a time scale of days,  weeks,
months and years;  or to a species development of linguistic abilities
(the
{\sc phylogenesis}
of language) on the time-scale appropriate to human evolution.
\paragraph{Sapir-Worf hypothesis.}\lb{sw}
A problem arises because concepts in one language might not have a direct
counterpart in the other language;  this is an aspect of the
Sapir-Whorf hypothesis,  Kay and Kempton (1984) \cite{bi:KK}.
If the Sapir-Whorf hypothesis is true then it suggests that meaning can evolve in isolation,
so that there are analogs of allopatric evolution,
if not then there are not.
\paragraph{Metaphor as Exaptation}\lb{me}
Exaptation has been defined in \S\ref{pe}.
Let the counterpart of the exapted parts be words,
or more accurately in the present context the meaning of words.
Words are given new uses other than their `original' use in metaphor.
For example in the metaphor `the mouth of a river',  mouth can be
considered to be an exapted word.   Of course `exapted word'
is now itself a metaphor: the point in making it is that it highlights
that in both biological evolution and in language its use is
both opportunistic and creative.
Both the metaphor and the original meaning can co-exist in language,
whereas a biological exaptation corresponds to only one characteristic,
and in this sense they differ.
\paragraph{Spandrels}\lb{sp}
In biological evolution spandrels are parts which originally occurred
as a secondary byproduct of primary structure.
Letting their counterpart of the spandrelized parts be words
is again an option,  but a preferable one is to let the counterparts
be phrases.   Idiomatic phrases perhaps occur as a by-product
of other structure,  but a better example is puns where
the secondary meaning of the pun can be thought of as a by-product of
the primary intention conveyed by the sentence.
\paragraph{Homologues}\lb{hm}
Homologous words certainly occur.
Camera,  chamber,  and comrade all have the same evolutionary
origin in the Greek word {\bf kamara},
but they now have completely distinct meanings.
\paragraph{Analogues}\lb{an}
Analogous words also occur.   Tetrads,  verbeins,  and basis vectors,
terms used respectively by relativists,  particle physicists,
and pure mathematicians mean much the same.
Examples of analogous words such as these can always be argued to differ in
nuance;  but the difference is slight and appears to be less that the
difference between the precise functions of standard biological analogues
such as the functions of fins of fishes and the paddles of whales.
\paragraph{Convergence}\lb{cg}
Convergence between word meaning frequently occurs,  for example between
transitory and caducity.   {\bf Caducity} originally meant a tendency to fall,
from the Latin
{\sc cadere}
= to fall.  ("The caducity of language,
in virtue of which every effusion of the human spirit is lodged in a body of death.",
attributed by the OED to M.Pattison (1813-84)).
Note also that a words {\it sounds} may be identical in two languages and yet have
completely different meanings,  depending on the language community of the listeners.
\paragraph{Susceptibility to Initial Conditions}\lb{si}
In natural languages the meaning of the two distinct words will never
converge to such an extent that they concur in meaning,
because they will always have different nuances.
An example of this is every time a
language users reads a script written before the words have closely converged.
Thus reticulate evolution between word meaning is not possible for natural
languages.   Now in a formal language it is possible to require by {\bf fiat}
that two words have identical meaning:  thus at first
sight reticulate evolution is possible for formal languages,
however the position is more complex than this.
Consider the definition of '{\sc derivative}' in mathematics,
Thurston (1994) \cite{bi:thurston} p.163.
There are many definitions possible,  Thurston lists as his first seven:  the
infinitesimal,  the symbolic,  the logical,  the geometric,  rate,
approximation,  and the microscopic, - and as about the 37th:
the cotangent bundle.   There is a choice as to which - if any - is considered
the primary definition and which are considered properties.   Now if in a
formal text two words are required to have the same meaning this can be
thought of as a reflection of the view of the writer,
experienced practitioners of the subject may have other preferences.
Thus reticulate evolution of a solitary word in a formal language
is only possible for an individual user of that formal language,
for the community which uses the formal
language reticulate evolution is not possible,
so that there is no susceptibility to initial conditions.

From the vantage point of an individual,
the meaning of a sentence obtained from
radical interpretation is unusual because
there is little sensitivity to initial conditions.
This is illustrated by people who are related and have been subject
to similar environments can have very different views,
it seems safe to say that their ideas on what words
and sentences mean will not vary substantially.
From the vantage point of a language community,  again there seems to be
little dependence on initial conditions:
for example language communities develop,
presumably on occasion independently words to describe the focal colours,
Berlin and Kay (1969) \cite{bi:BK}.
Susceptibility to initial conditions is important because of the role
that initial conditions hold in dynamical theory.
Dynamical systems theory is the mathematical theory which it is hoped will
eventually be used to describe all time dependent phenomena.
For its application to biological evolution see for example
Hofbauer and Sigmut (1988) \cite{bi:HS},
its terms have been used by analogy to describe
individual word meaning see for example
Hinton and Shallice (1991) \cite{bi:HinS},
and for its wider application in theories of meaning see
Hoffman (1987) \cite{bi:hoffman}.
\paragraph{Commensurability}\lb{cc}
Kuhn (1970) \cite{bi:kuhn} and Ramberg (1990) \cite{bi:ramberg} Ch.9 discuss
the concept of incommensurability,
a simple definition of this is that the usage of a word becomes
{\sc incommensurable}
when its meaning has substantially diverged in separate language communities.
The definition begs the question of how substantially different
and separate affairs have to be before it is allowable to apply the word incommensurable.
If one considers the counterpart of a word to be a species then
this is a difference from biological evolution
where a naive view is that it is in principle,  but perhaps not in practice,
always possible to tell if a speciation has occurred,
for example by seeing if fertile offspring can be produced from two strands.
This naive view of speciation does not always hold,
see the remarks on reticulate evolution at the end of \S3.1,
the difference is more one of degree rather than an absolute,
there can be fusion between species.
The essence of the above distinction is that words become
incommensurable gradually,  whereas,  in principle at least,
a species has either split or it has not.
The situation becomes more complex if one considers the counterpart
of a word to be other aspects of biological structure.
For example if the counterpart to a word is taken to a gene,
then again the gene has either changed or not,
there is no gradual change of use as in incommensurable words;
however whether the effects of an unchanged gene can change gradually is less clear.
\paragraph{No Counterpart to Speciation}\lb{cs}
In biological evolution speciation is the evolution of a new biological species.
It is not clear what should be taken to be the counterpart of a species in language.
The most common view is that species are the analogs of languages,
geographical races are the analogs of dialects,
and organisms are the analogs of idiolects.
In this case the comparison is between a species and an individual language,
as discussed above a species has either split or it has not,
whereas languages merge from on to another,
so that viewed as a time-dependent process they behave differently.
Schleicher took the view that languages are organisms,
then a comparison of the nature of time-dependent process is then even further apart.
A noeme might appear to be a good counterpart to a species;
the reason this does not work is,  again,
that there are fairly clear criterion for when a new species has been formed,
and there is no such criteria for when a new glosseme has formed.
One needs a linguistic structure that changes in a more discrete manner
for there to be a good counterpart.
The only place where linguistic structure changes in such a discrete manner is in formal
languages,  where is possible to say that a definition is a counterpart of a species.
The nature of their time-dependent change might be similar,
but it is not clear how other structure in biology and formal languages would correspond,
for example there is nothing corresponding to a theorem.
\paragraph{No Counterpart to the Traditional Gene}\lb{tg}
For the purpose of our comparison we use the traditional naive definition
of a gene,  here  called the traditional gene,
as a minimal unit of inheritance of definite size and
{\it information content}.
The reason that a traditional gene cannot be equated with a word,
is because of the fixed information content of a traditional gene;
a words information content can alter in natural language,
although it could be fixed in a formal language or a computational model.
The traditional gene might be a convenient concept,
but the concept of a gene has now changed.
Until 1945 the definition of a gene as a unit of inheritance was sufficient
to cover all known genetic data,  Lewin (1994) \cite{bi:lewin} p.70;
however developments since then have
shown this naive definition to be inadequate.
The idea that a gene has a definite size is in
fact incorrect,  genes can disperse from their original cluster on a
chromosome to new locations by chromosomal rearrangement,
Lewin (1994) \cite{bi:lewin} p.731;  thus a gene can no longer be
thought of as a unit on a chromosome as it has been spread around.
An example of indefinite gene size is given by the gene for fragile {\bf X}
syndrome which results in mental retardation in boys.
This gene contains repeated sequence  of nucleotides which can expand
or contract.   If there are relatively few repeats (say less than 20)
then fragile {\bf X} syndrome does not occur;
however if many repeats occur then the syndrome occurs.
The information content of genes is not fixed either.
Genes vary and are becoming increasingly difficult to define.
In the 1940's Beadle and Tatum proposed `one gene for one protein (enzyme)'.
When multi-subunit enzymes were discovered this definition was modified
to `One gene for one polypeptide.'.
Then genes were found to code for RNA molecules that never were
translated into proteins.   Even if one just considers the genes that
specify proteins,  it is now
clear that some genes code for just part of a polypeptide chain
(e.g. in the case of antibodies)
while others code for more than one polypeptide.
In one tissue a particular stretch of DNA (a gene,  on the old view)
will code for one polypeptide while another time it might
code for a different (albeit related) polypeptide of a different length.
Finally,  it is not possible
these days even to predict with total accuracy what amino acids
in the polypeptide will be from the sequence of the ``gene''.
This is because editing of the mRNA can alter the information
content of the gene,  as {\bf Figure 1} illustrates,  with the '$*$'
indicating the edited mRNA:\\

\begin{figure}
   \centering
   {\epsfig{figure=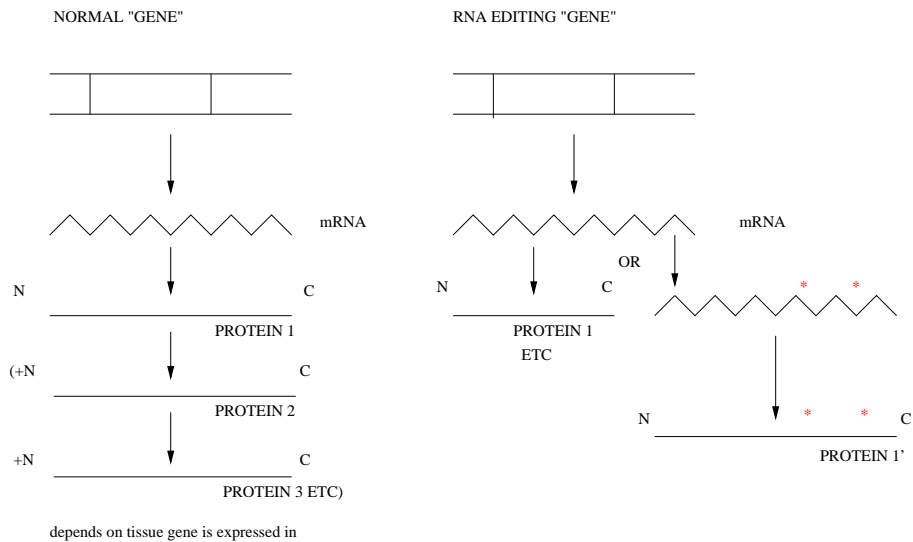,height=2.8in}}              
   \caption{Editing of mRNA}
\end{figure}

Many textbooks just say that a gene is ``a minimum unit of inheritance''
but do not dare to say more as there are no further general rules.
Even to say that genes are composed of DNA or RNA is probably inaccurate.
Evidence for transmissible infectious agents
that lack nucleic acids (Prions) is accumulating,
see for example Lewin (1994) \cite{bi:lewin}.
It may well happen that new concepts {\it emerge}
from these consideration that suffice to explain the modern
data and reduce to the concept of a traditional gene in some limit.
It is curious to note that molecular biologists
use the concept of words and languages frequently,
for example gene libraries and dictionaries.
For a modern gene a comparison with language
is more complex as modern genes can have different functions.
One could equate a modern gene with a word and
perhaps say that words also have different function depending
on their syntactic group,  noun,  verb, etc\dots.
\paragraph{No Counterpart to Hox Genes.}\lb{hox}
{\sc Hox genes} are involved in the control of development of structures
that display iterative homology,  McGinnis and Krumlauf \cite{MK} (1992).
The key part of the definition is 'iterative',
the only thing in natural language that suggests an iterative process is at work
is language acquisition.
Large amounts of syntax are learned very quickly by children,
and this suggests that there is some cognitive structure which can be
applied iteratively to do the learning.
The problem with identifying some as yet unknown cognitive structure
with the Hox gene is that the Hox gene is a particular type of gene,
and genes perhaps correspond best to words,  rather than acquisitional structure;
so that there would seem to be no counterpart to Hox genes in language.
Another way of looking at a comparison is to note that Hox genes cause
a lot of genes to move together, so that the counterpart of Hox genes would be
related to semantic clusters, Agarwal \cite{agarwal} (1995),  moving together.
There are examples of concepts moving together when there is
a scientific or technical advance.
A mathematical example is topology where many concepts originate in real analysis.
In physics there are examples of many concepts transferring,
say in newtonian mechanics from its original formulation to its Hamiltonian formulation,
however there are many cases where this does not happen.
Take for example 'force' in newtonian mechanics,
in vacuum general relativity the concept simply does not exist
as particles move along geodesics;
how to transfer other concepts,  such as angular momentum,  causes problems.
A technical example would be the concepts used to describe aircraft,
many of which originated in the description of ships.
Whether there is an analog between the theories
seems to depend on what property of Hox genes one wants to emphasize.
\paragraph{Compositionality}\lb{cm}
Once we have limited ourselves to traditional genes then it follows
that memes,  noemes,  and similar constructions are fundamentally different
from traditional genes,  because traditional genes have a definite size
and information content but the size of memes and noemes is indefinite.
To illustrate the indefinite size of these objects begin by recalling
the definition of a correspondence theory given at the beginning of \S\ref{sec:comparison}.
Next define {\sc compositionality} as the term used
to say that the meaning of a sentence depends on the meaning of its parts,
see for example Zadrozny (2000) \cite{bi:zad}.
This definition implies a restricted domain of applicability for a noeme.
Compositionality must be at best only an approximation,
because it does not take into account environment or context.
This has been noted by philosophers for example Davidson (1984) \cite{bi:davidson2} p.22
\begin{quote}
Frege said that only in the context of a sentence does a word have meaning;
in  the same vein he might have added that only in the context of a language
does a sentence (and hence a word) have meaning.
\end{quote}
One could take the view that the traditional gene has the property of compositionality.
\paragraph{The Antinomy of the Liar}\lb{al}
It could be added that language itself is a changing structure,
dependent on many aspects of the world,
and that these should also be taken into account in
order to say what meaning is;
indeed it is not clear {\it a priori} what information could be
excluded from a theory of meaning,
and this implies that there can never be an absolute
correspondence between language and the world.
When considering the amount of
language necessary for a truth it is straightforward to show that
sentences are insufficient by
constructing self-referential truth ({\bf SRT}) sentences,
such as the
{\it antinomy of the liar}
an example of which is `what this sentence says is not true'.
Now the way in which the truth of
such sentences can be assigned is by taking into account sufficient
context to disambiguate their self-referential nature.
However the amount of language necessary for as theory of
meaning cannot be disambiguated in such a manner,
as the self-referential meaning ({\bf SRM}) sentences:
`This sentence is meaningless',
or `What this sentence means is the opposite of what you think it means.',
and the hybrid self-referential truth/meaning ({\bf SRH}) sentence:
`If this sentences is true then it is meaningless',  show.
The above suggests that meaning is more context dependent than truth.
\paragraph{Prefix Meaning}\lb{pm}
Appealing to context is a vague procedure
without elucidating the nature of its component factors.
On the other hand meaning can be
assigned to small linguistic segments such as prefixes and affixes,
for example {\em bi}-,  {\em trans}-,  etc\dots,  which
certainly have some meaning associated with themselves,
as for example given by the normative definitions of a dictionary.
A {\sc codon} is a unit that consists of three adjacent bases on a DNA molecule
and that determines the position of a specific amino acid in a protein molecule
during protein synthesis.
The counterpart of a codon could be a declension or a grammatical inflection.
A codon could be both the counterpart to that and also to affix meaning,
the ''s'' denoting plural,  is both an affix and a decliner.
Another analog of prefix meaning is
{\it content addressable memory},
which is the ability of people and some machine programs to retrieve
information based on its
contents rather than simply from an address.
The more information which is available the
easier it is to retrieve the rest:
analogously for meaning,  the greater the linguistic structure
and context present the easier it is to assign meaning.
A formal concept which perhaps
encompasses both cases is
{\sc distal access},
Touretzky (1990) \cite{bi:touretzky},
which is the ability to reference a
complex structure via an abbreviated tag.
Thus when it comes to assigning meaning to language the size of structure
that this happens to is unclear.
There appears to be nothing analogous a traditional gene;
there is no unit between part of a word and a whole
language that it is possible to say encapsulates meaning.
Molecular biology shows the inadequacy of the traditional idea of a
gene as having definite size and information content,
thus the difference is more one of degree rather than an absolute,
however the absence of a unit is much more severe in language.
\section{Conclusion}
\label{sec:sec7}
To  summarize it is sometimes said that we lean by approximation:
radical interpretation is a prescientific theory which
hopes to describe and perhaps ultimately model how,
by iterative approximation,  we learn what meaning is.
It is prescientific in the sense that it is not well enough developed to
say what data would refute it;
also there appears to be no alternative theory to judge against it.
If a correspondence theory of language is assumed then radical interpretation
is a process by which mental representations of the world become more accurate.
Here it has been
argued that radical interpretation applies not just to sentences,
but to all socially contrived non-physical structure;
in particular it applies to noemes,  memes,  and conventions.
It was argued that radical interpretation is a typical time dependent
process with the properties that,
subject to certain qualifications explained in the text:
on the fine-grained time scale it
proceeds by punctuated equilibrium but on the coarse-grained
time scale it proceeds gradually.
Exaptation occurs, an example being metaphor.  Spandrels,  homologues,
analogues,  and convergence also occur.
Reticulate evolution can only occur in a formal
language for an individual user where it can be introduced by fiat;
this cannot happen for a
language community using a formal language or for natural languages.
Radical interpretation shows little sensitivity to initial conditions.
Viewed as a time dependent process radical
interpretation has more differences than similarities
with biological evolution;  the origin of this
is that (for the purposes of pre-1945 biology at least)
a convenient unit,  with a definite information content,  the traditional gene,
exists in biology,  but no such unit can be found in a theory of meaning.
\section {Acknowledgements}
I would like to thank Drs.C.D.O'Connor and W.A.C.Mier-Jedrzejowicz
for their interest in this paper,
and the editor and reviewers of {\it Behavior and Philosophy}
for their many comments.

\end{document}